%% file: main.tex
\documentclass[letterpaper]{article} 
\usepackage{aaai24}  
\usepackage{times}  
\usepackage{helvet}  
\usepackage{courier}  
\usepackage[hyphens]{url}  
\usepackage{graphicx} 
\usepackage{natbib}  
\usepackage{caption} 
\usepackage{algorithm}
\usepackage{algorithmic}

\usepackage{newfloat}
\usepackage{listings}
\DeclareCaptionStyle{ruled}{labelfont=normalfont,labelsep=colon,strut=off} 
\floatstyle{ruled}
\newfloat{listing}{tb}{lst}{}
\floatname{listing}{Listing}
\usepackage{hyperref}
\hypersetup{hidelinks,
	colorlinks=true,
	allcolors=blue,
	pdfstartview=Fit,
	breaklinks=true}

\title{Self-Supervised Bird's Eye View Motion Prediction with Cross-Modality Signals}
\author {
    Shaoheng Fang\textsuperscript{\rm 1},
    Zuhong Liu\textsuperscript{\rm 1},
    Mingyu Wang\textsuperscript{\rm 2},
    Chenxin Xu\textsuperscript{\rm 1},
    Yiqi Zhong\textsuperscript{\rm 3},
    Siheng Chen\textsuperscript{\rm 1,4}
}
\affiliations {
    \textsuperscript{\rm 1} Shanghai Jiao Tong University \\
    \textsuperscript{\rm 2} University of Chinese Academy of Sciences \\
    \textsuperscript{\rm 3}  University of Southern California \\
    \textsuperscript{\rm 4} Shanghai AI Laboratory\\
    \{shfang, xcxwakaka, sihengc\}@sjtu.edu.cn, zuhong.liu@polytechnique.edu, \\
    wangmingyu21a@mails.ucas.ac.cn, yiqizhon@usc.edu
}

\usepackage{bibentry}

\usepackage{xcolor}

\usepackage{multirow}
\usepackage{booktabs}
\usepackage{amsfonts}
\usepackage{makecell}
\usepackage{amsmath}

\begin{document}

\maketitle

\input{contents/0-abstract}
\input{contents/1-introduction}
\input{contents/2-related_works}
\input{contents/3-method}
\input{contents/4-experiments}
\input{contents/5-conclusion}

\clearpage

\input{contents/acknowledgement}

\bibliography{aaai24}

\input{contents/supp}

\end{document}

%% file: contents/0-abstract.tex
\begin{abstract}
Learning the dense bird's eye view (BEV) motion flow in a self-supervised manner is an emerging research for robotics and autonomous driving.
Current self-supervised methods mainly rely on point correspondences between point clouds, which may introduce the problems of fake flow and inconsistency, hindering the model’s ability to learn accurate and realistic motion.
In this paper, we introduce a novel cross-modality self-supervised training framework that effectively addresses these issues by leveraging multi-modality data to obtain supervision signals.
We design three innovative supervision signals to preserve the inherent properties of scene motion, including the masked Chamfer distance loss, the piecewise rigidity loss, and the temporal consistency loss.
Through extensive experiments, we demonstrate that our proposed self-supervised framework outperforms all previous self-supervision methods for the motion prediction task. Code is available at \href{https://github.com/bshfang/self-supervised-motion}{https://github.com/bshfang/self-supervised-motion}.
\end{abstract}

%% file: contents/1-introduction.tex
\section{Introduction}

Accurate prediction of dynamic motion within a scene is fundamental for the safe and robust planning of autonomous vehicles. Instead of predicting instance-level trajectories~\cite{chen20203d}. An emerging trend is to predict the dense motion flow in the BEV (Bird’s Eye View) map directly from raw sequential sensor input in an end-to-end manner~\cite{wu2020motionnet, besti}. This approach is less susceptible to perception errors and possesses the capability to discern class-agnostic motion~\cite{wu2020motionnet, wong2020identifying}. 
Nevertheless, training flow prediction models with supervision necessitates a substantial volume of annotations for sensor data and annotating motion labels for sensor data proves to be intricate and costly. Hence, the effective utilization of vast amounts of unlabeled raw data for motion prediction training has emerged as a notable and encouraging challenge.
Recently, many works have proposed various self-supervised frameworks to learn the BEV motion without relying on ground truth labels~\cite{pillarmotion,weakmotionnet,jia2023contrastmotion}.

Inspired by self-supervised scene flow estimation, current self-supervised BEV motion prediction methods~\cite{pillarmotion, weakmotionnet} primarily rely on chamfer distance loss to establish the point-level correspondences between point clouds.
However, this heavy dependence on point-level correspondences leads to two major problems when learning motion patterns from real-world LiDAR point cloud data.

\begin{figure}[t]
\centering
\includegraphics[width=0.95\columnwidth]{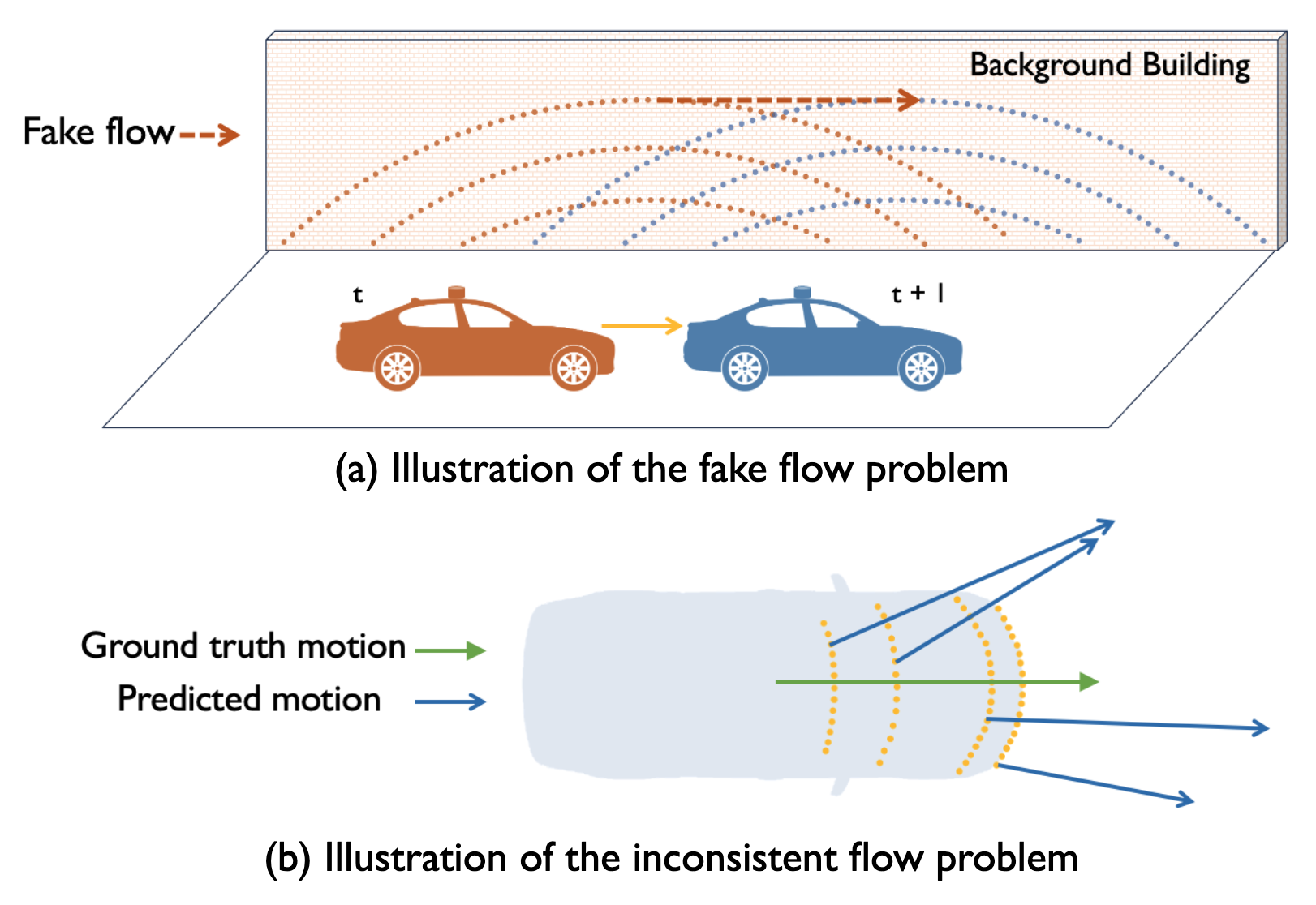}
\vspace{-2mm}
\caption{Problems in current self-supervised motion learning methods that rely on point correspondence. (a) For static objects (background building), points with correspondences in the point cloud sequence may have completely different locations, misleading the model to learn the fake flow. (b) Due to the sparse nature of the point cloud, points within an instance may learn highly varying flow.}
\label{fig:optf_and_oversegment}
\vspace{-4mm}
\end{figure}

The first problem is fake flows. 
Due to the alterations in the viewpoint of the LiDAR sensor, points associated with the background or static objects often exhibit flow that does not exist, as shown in Figure \ref{fig:optf_and_oversegment}(a). 
This fake flow will mislead the model to learn incorrect motion patterns, thereby adversely affecting the accuracy of predictions. One previous work~\cite{weakmotionnet} introduces a weakly supervised setting where foreground/background ground truth is available to mitigate the impact of noise originating from background points to alleviate the problem. However, the method still remains limited as extra human annotation is indispensable.

The second problem is the inconsistent flows within one single object; see an illustration in Figure \ref{fig:optf_and_oversegment}(b). Owing to the inherent sparsity in point cloud data, the point-level flows associated with the same objects may exhibit inconsistent motions when solely relying on the point correspondences. This problem of inconsistency violates the object-level rigid constraints and causes confusion for model learning procedures. \cite{pillarmotion} aims to preserve the local uniformity of motion flow by employing a smoothness loss, which encourages minimal changes among neighbor flow values. Unfortunately, this assumption fails in the boundary region between moving objects and the background and is unable to ensure instance-level motion consistency.

To address the challenges of fake flow and inconsistent flow, rather than developing another network model, our focus is to design dedicated supervision signals to preserve a series of inherent properties of scene motion. For the fake flow issue, one common, yet fundamental property is that motion is restricted to moving objects. Stationary components should exhibit no motion, allowing us to filter out background noise and obtain a more precise motion flow. To tackle the inconsistent flow challenge, we focus on two primary properties: object and temporal consistency. In essence, points within rigid objects should move uniformly. Furthermore, object motions should remain relatively stable over short periods, ensuring no abrupt changes. By emphasizing these key properties, we enhance the uniformity and reliability of the motion flow.

However, due to the notorious noise and sparsity issues of point cloud data, relying solely on point cloud sequences might compromise the accurate representation of scene motion properties. To compensate, we leverage the multi-modality information. We specifically incorporate sequential camera images—readily accessible since most robots are equipped with cameras, thereby incurring no extra annotation costs. These sequential images enable the extraction of optical flow, providing a rich layer of motion insights. This stands in stark contrast to the sparse, irregular, and fragmented data in point cloud sequences. Optical flow images distinctly highlight the coherent and consistent motion of objects, sharply separating them from the background.

Leveraging the advantages of multi-modality data, we incorporate the spirit of preserving scene motion's inherent properties into a novel self-supervised training framework for BEV motion prediction. Specifically, i) to ensure that motion is exclusive to moving objects, the framework generates a pseudo static/dynamic mask for each point cloud according to the optical flow data. Then this mask will be used to ensure structural consistency exclusively for the dynamic portion through a novel masked Chamfer distance loss; ii) to promote motion consistency in individual objects, we employ a simple clustering technique to the optical flow image, discerning instance boundaries and creating pixel clusters in the image space. Then the cluster information for each pixel will be projected to the point cloud space, creating rigid point cloud clusters that should share the same motion flow to ensure the instance-level rigidity constraints; and iii) for temporal motion consistency, we introduce a novel temporal consistency loss, which enforces the smoothness of predictions across long point cloud sequences. Note that image data are only used for providing supervision signals in the training phase; during inference, the proposed BEV motion prediction network only needs point cloud sequences.

Experimental evaluations conducted on the nuScenes~\cite{caesar2020nuscenes} dataset demonstrate that our proposed methodology improves upon previous self-supervised approaches by up to $40\%$. Notably, our method achieves performance comparable to weakly-supervised and fully-supervised methods.

To summarize, the main contributions of our work are:
\begin{itemize}
    \item We propose a novel cross-modality self-supervised training framework for BEV motion prediction, which leverages multi-modality data to obtain supervision signals.

    \item We propose three novel supervision signals to preserve the inherent properties of scene motion, including the masked Chamfer distance loss, the piecewise rigidity loss and the temporal consistency loss.
    
    \item Our method achieves state-of-the-art performance. Comprehensive experiments demonstrate the effectiveness of our designed framework.
\end{itemize}

%% file: contents/2-related_works.tex
\section{Related Work}

\begin{figure*}[ht]
\centering
\includegraphics[width=0.95\textwidth]{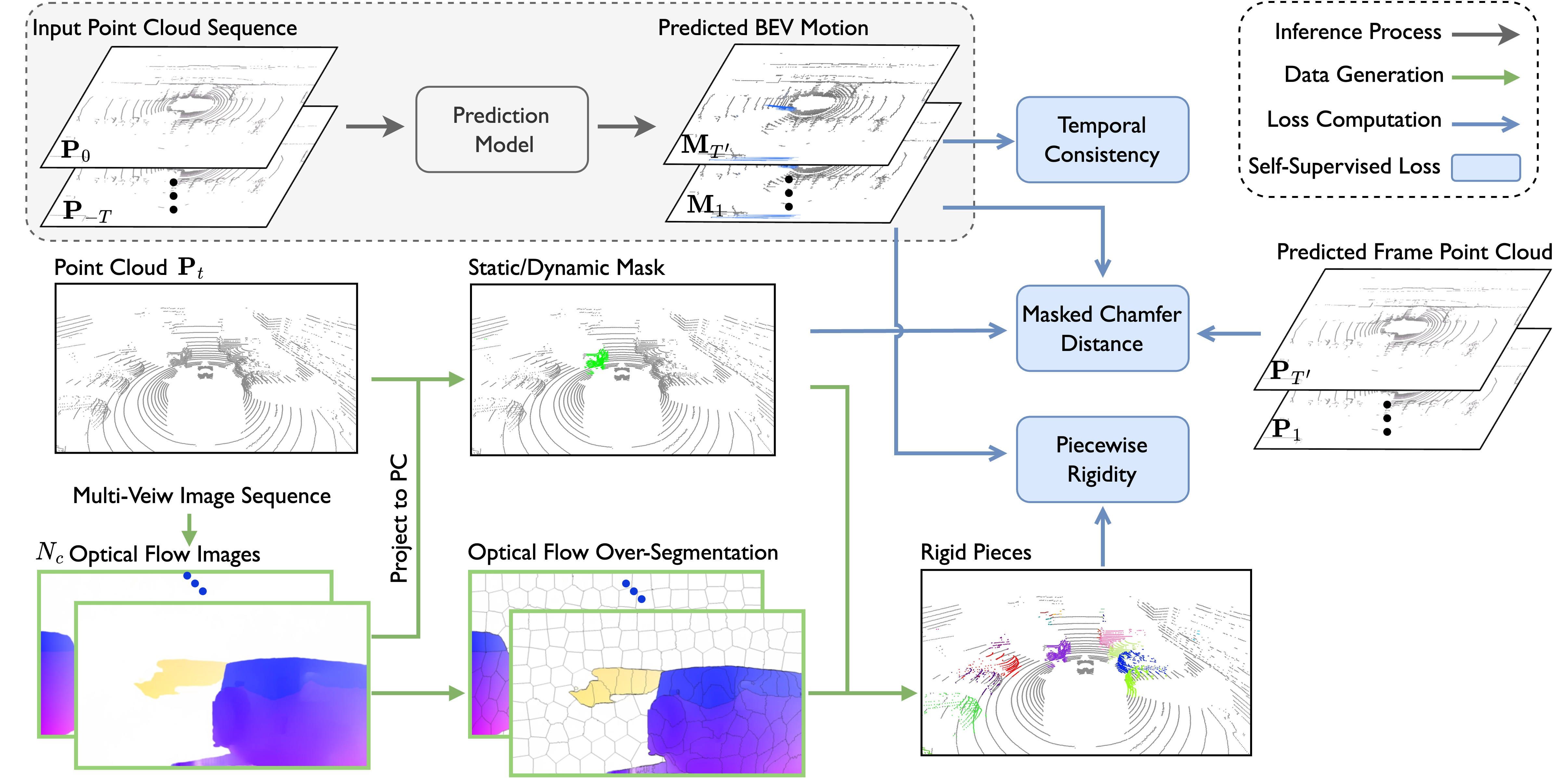}
\caption{An overview of our cross-modality self-supervision learning framework. An overview of our cross-modality self-supervision learning framework. For self-supervised training, we introduce three innovative self-supervised losses that align with real-world motion patterns.  The inference process only takes the point cloud sequence as input and predicts the motion flow of each BEV cell (grey area). }
\label{fig:overview}
\vspace{-3mm}
\end{figure*}

\subsection{Motion Prediction}

The goal of motion prediction is to estimate the future movements of mobile objects in a scene based on past observations. Traditional approaches tackle this issue via a two-stage framework, relying on the results of 3D object detection and tracking to predict the instance-level trajectories~\cite{casas2020spagnn, luo2018faf, phillips2021deep}. However, the dependence on intermediate results may lead to error accumulation and a limited ability to perceive unknown classes~\cite{wu2020motionnet, wong2020identifying}. An emerging trend is to predict dense future motion in an end-to-end framework directly from sequential sensor input, including multi-frame point clouds~\cite{wu2020motionnet, lee2020pillarflow, pillarmotion, filatov2020any, besti, wei2023asynchrony} and multi-view images~\cite{hu2021fiery, zhang2022beverse, fang2023tbp}.

Training a motion prediction model requires high-quality manual labels, but obtaining such labels is both expensive and laborious. Accordingly, some methods aim to mitigate this issue from various perspectives. \cite{pillarmotion} proposes a self-supervision method that utilizes point cloud structure consistency and cross-modality regularization; \cite{weakmotionnet} proposes the use of a weakly supervised setting that only utilizes foreground/background information, effectively improving the accuracy. \cite{jia2023contrastmotion} employs contrastive learning to learn BEV pillar features and uses pillar association to predict motion. In this paper, we propose a novel self-supervised framework and achieve remarkable performance that is comparable to other weakly supervised and even fully supervised approaches.

\vspace{-1mm}
\subsection{Self-Supervised Scene Flow Estimation}

Scene flow estimation aims to determine the 3D motion displacement at the point level between a pair of point clouds~\cite{liu2019flownet3d, puy2020flot, jund2021scalable, cheng2022bi, li2021hcrf, gu2019hplflownet}. Learning scene flow in a self-supervised manner is a popular field of research~\cite{mittal2020just,wu2019pointpwc,baur2021slim,kittenplon2021flowstep3d,tishchenko2020self}. ~\cite{mittal2020just} was the first to establish a self-supervised learning framework that utilizes a combination of nearest neighbor and cycle consistency loss. Following ~\cite{wu2019pointpwc},  \cite{kittenplon2021flowstep3d,pontes2020scene} use the chamfer distance loss to learn the point correspondences between two point clouds. \cite{li2022rigidflow, gojcic2021weakly} employ ego-motion estimation and exploit the piecewise rigid nature of point clouds.

We follow the philosophy of \cite{wu2019pointpwc, li2022rigidflow} by designing self-supervised loss to maintain structural consistency between point clouds and exploiting the piecewise rigidity for regularization.
Nevertheless, most of these methods~\cite{li2022rigidflow,wu2019pointpwc,mittal2020just} usually assume strong one-to-one correspondences between point clouds and incur heavy computational costs, making them unsuitable for real-time perception in autonomous driving. We propose a masked chamfer loss to mitigate these issues.
Moreover, unlike scene flow estimation, which identifies motion between a pair of point clouds, we concentrate on predicting the future of the scene based on point cloud sequences.

\subsection{LiDAR-Camera Fusion}

LiDAR-camera fusion has been extensively investigated to enhance scene perception, including various tasks such as 3D object detection~\cite{vora2020pointpainting, li2022deepfusion, liang2022bevfusion} and scene flow estimation~\cite{rishav2020deeplidarflow, liu2022camliflow}.
A novel line of research is to leverage cross-modality information as supervised signals to support model training. \cite{ding2023hidden} combines detection and tracking results from LiDAR point clouds with odometry data and optical flow to jointly improve radar scene flow learning. \cite{li2022learning_noisy} generates noisy pseudo-labels from optical flow to supervise scene flow learning. Additionally, \cite{pillarmotion} facilitates motion prediction learning through LiDAR-camera cross-modality regularization. Optical flow data, which can be easily obtained from camera video without human labeling, has shown the potential to aid motion learning on point clouds.
However, these methods solely employ the numerical values of optical flow as the guidance for point cloud motion and ignore the inherent advantages of optical flow data over point clouds.

%% file: contents/3-method.tex
\vspace{-1mm}
\section{Method}
\label{sec:method}

This section introduces a self-supervised training framework for BEV motion prediction, where three novel supervision signals are generated from multi-modality inputs, including point cloud sequences and camera videos.

\begin{figure*}[t]
\centering
\includegraphics[width=0.95\textwidth]{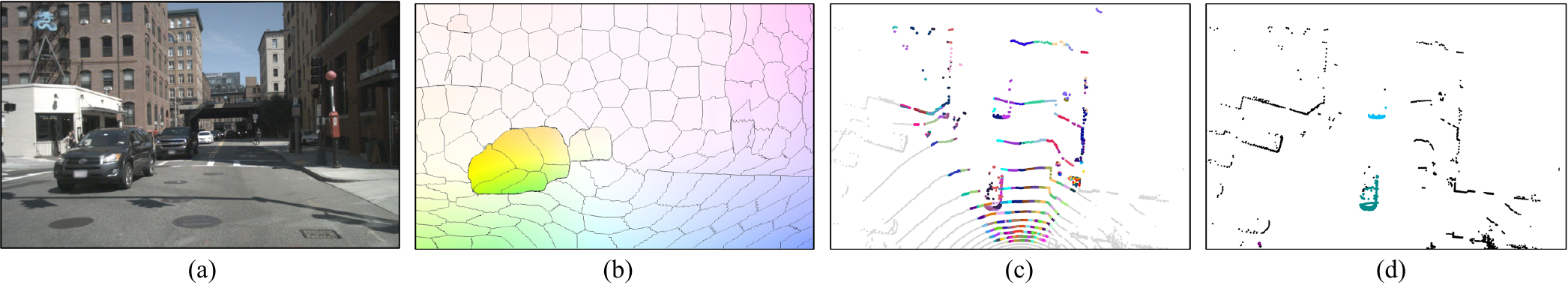}
\vspace{-2mm}
\caption{Rigid piece generation. (a) A frame of sequential images; (b) Over-segmentation on the optical flow image; (c) Over-segmentation projected to the associated point cloud; (d) Rigid pieces after fusion. In (c) and (d), each color refers to a piece.}
\label{fig:piecewise_rigidity}
\vspace{-3mm}
\end{figure*}

\subsection{Problem Formulation}

The objective of the motion prediction task is to directly forecast the motion of mobile grids in the 3D BEV map from historical point cloud sequences~\cite{weakmotionnet, besti}. The prediction model takes the current frame $0$ along with $T$ past frames of point clouds that synchronized to the current frame as input. The point cloud sequence is denoted as $\mathbf{P}_t=\{ p_i^t \in \mathbb{R}^3 \}_{i=1}^{N_t}, ~t=0,-1,\cdots,-T$, where $N_t$ represents the number of points in $P_t$. The multi-view camera video is utilized in the training process. The corresponding multi-view images of $\mathbf{P}_t$ are $\{ \mathbf{I}_t^{k} \in \mathbb{R}^{H \times W \times 3} \}_{k=1}^{N_c}$, where $N_c$ is the number of cameras.

The future motion is represented in the form of a BEV (bird's eye view) map. Assuming the model predicts $T^{\prime}$ frames of future motion field, $\mathbf{M}_t \in \mathbb{R}^{X \times Y \times 2}, ~t=1,\cdots, T^{\prime}$ represents the motion field of the $t$ frame, where $X \times Y$ is the shape of the BEV map according to the vehicle-ego coordinates at the current timestamp.
Considering that each grid in the BEV map represents a rather small area in the real-world scene, points within the same pixel grid have identical motion flow. 
To generate a point-level 3D motion flow, each point can be assigned the motion of its corresponding position in the BEV map based on its 3D coordinates, and the vertical motion is set as zero. The motion of points is denoted as $\mathbf{F}_t = \left\{ f_i^t \in \mathbb{R}^3 \right\}_{i=1}^{N_0}, ~t=1,\cdots, T^{\prime}$.

Apart from predicting the future motion from time 1 to $T^{\prime}$, the model can also infer the motion situation from the current frame to the past frames. Therefore, in the following section, $\mathbf{T} = \{t_1, \cdots, t_n\}$ is used to represent the entire set of time frames that the model predicts, including $1$ to $T^{\prime}$ and possible backward predictions (for example, $-1$).

\subsection{Overview}
Figure \ref{fig:overview} overviews our training framework. Since the prediction model is not our focus, we directly adopt MotionNet~\cite{wu2020motionnet}. The key of this work is to leverage multi-modality inputs to provide three supervision signals that can preserve inherent properties of scene motion. They include: 1) pseudo static/dynamic mask loss generated from sequential video, 2) the piece-wise rigidity loss, and 3) temporal motion consistency loss. Here are the detailed descriptions.

\subsection{Pseudo Static/Dynamic Mask Loss}

We employ Chamfer Distance as the foundation of learning structural consistency. The Chamfer Distance serves as a measure of similarity between two sets of points. In self-supervised training~\cite{wu2019pointpwc, kittenplon2021flowstep3d} or optimization~\cite{li2021neural, pontes2020scene} methods related to point clouds, the chamfer distance loss is a widely used technique that helps maintain the structural consistency of two point clouds. 

For any frame $t$, with the predicted point-level flow $\mathbf{F}_t$ from frame $0$ to frame $t$, the predicted point cloud can be calculated as $\mathbf{P}^{\prime}_t = \{ p_i^{\prime} \in \mathbb{R}^3 \mid p_i^{\prime} = p_i^0 + f_i^t \}_{i=1}^{N_0}$.
Given the point cloud $P_t$ at frame $t$, the self-supervised chamfer distance loss can be defined as 

\begin{small}
\begin{equation}
    \mathcal{L}_{cd}(\mathbf{P}_t, \mathbf{P}^{\prime}_t) = \sum_{p_j \in \mathbf{P}_t} \min_{p^{\prime}_i \in \mathbf{P}^{\prime}_t} \left\| p_j - p^{\prime}_i \right\|_2^2 + \sum_{p^{\prime}_i \in \mathbf{P}^{\prime}_t} \min_{p_j \in \mathbf{P}_t} \left\| p^{\prime}_i - p_j \right\|_2^2.
\label{eq:cd_loss}
\end{equation}
\end{small}

However, the point cloud data is often sparse and full of noise points. Even for stationary objects, the point cloud representation can vary significantly with the sensor's movement~\cite{khurana2023point}. This poses great challenges and introduces noise when relying on Chamfer distance loss for learning. To better understand the motion of a dynamic 3D scene, it is crucial to focus on moving targets while disregarding the background and stationary objects. However, due to the sparse and noisy nature of the point cloud, it is difficult to distinguish between the static and dynamic parts of a point cloud in open scenes. In contrast, optical flow in the image space is much more accessible and easier to obtain. Video data is abundant with superior temporal and texture information, and the relevant techniques are already well-established~\cite{sun2018pwc,teed2020raft}. Previous works~\cite{pillarmotion, ding2023hidden} have utilized the value of image optical flow to assist in learning point cloud scene flow. In our method, we propose to extract a pseudo static/dynamic mask from the optical flow results of the image data to aid in structure consistency learning.

Given the point cloud time frame $t \in \mathbf{T}$, we can get the adjacent image pairs $(\mathbf{I}^k_t, \mathbf{I}^k_{t+\delta t}),~k=1, \cdots, N_c$ from the camera video. For brevity, we omit the superscript $k$ for camera index and the subscript $t$ for frame index in subsequent contents, and use $\mathbf{I}$ and $\mathbf{I}^{\prime}$ to denote $\mathbf{I}^k_t$ and $\mathbf{I}^k_{t+\delta t}$. The optical flow generated from $\mathbf{I}$ and $\mathbf{I}^{\prime}$ is denoted as $\mathbf{F}^{\text{2D}} \in \mathbb{R}^{H \times W \times 2}$.

The optical flow $\mathbf{F}^{\text{2D}}$ cannot yet be directly used to determine the motion status of each pixel. 
Apart from the optical flow generated by dynamic targets in the scene, the movement of the ego vehicle also produces flow in the camera view. Following ~\cite{pillarmotion}, we divide the optical flow into two parts, $\mathbf{F}^{\text{2D}} = \mathbf{F}^{\text{2D}}_{ego} + \mathbf{F}^{\text{2D}}_{mot}$, where $\mathbf{F}^{\text{2D}}_{ego}$ corresponds to the optical flow caused by vehicle motion and $\mathbf{F}^{\text{2D}}_{mot}$ corresponds to the optical flow caused by dynamic objects.

The numerical value of $\mathbf{F}^{\text{2D}}_{ego}$ can be calculated through the sensors' poses. Let $p_i \in \mathbf{P}$ represent a point within the image $\mathbf{I}$ and $\mathcal{T}_{\mathbf{P} \rightarrow \mathbf{I}}$ represent the transformation matrix from the lidar point cloud $\mathbf{P}$ to the image $\mathbf{I}$
\begin{equation}
    (u_i, v_i) = \mathcal{T}_{\mathbf{P} \rightarrow \mathbf{I}} (p_i).
\end{equation}
The value of $\mathbf{F}^{\text{2D}}_{ego}$ corresponding to $p_i$ is then
\begin{equation}
    \mathbf{F}^{\text{2D}}_{ego} (u_i,v_i) = \mathcal{T}_{\mathbf{P} \rightarrow \mathbf{I}^{\prime}} (p_i) - \mathcal{T}_{\mathbf{P} \rightarrow \mathbf{I}} (p_i).
\end{equation}

Ideally, the calculated $\mathbf{F}^{\text{2D}}_{mot}$ corresponding to a stationary target or background would be close to 0. Thus, static points can be distinguished by setting a small threshold for $\mathbf{F}^{\text{2D}}_{mot}$.

Nevertheless, when dealing with distant moving objects that are far from the camera, their corresponding optical flow values may be small and incorrectly classified as static. To mitigate such effect, we employ the projected 3D scene flow to supplement the static assessment.
Denote $\mathbf{F}^{\text{2D}}_{mot} (u_i,v_i)$ as $f_{i}^{\text{2D}}$. With the constraint of zero vertical motion, we can project the 2D optical flow $f_{i}^{\text{2D}}$ to a 3D scene flow originating from $p_i$. 
The operation is represented by a projection $\mathcal{T}_{\text{optf} \rightarrow \text{sf}}$ (see more info in supp.).
\begin{equation}
    f_{i}^{\text{3D}} = \mathcal{T}_{\text{optf} \rightarrow \text{sf}} (f_{i}^{\text{2D}}).
\end{equation}

The pseudo static/dynamic status $s_i$ of $p_i$ is estimated as 
\begin{equation}
s_i = \left\{
\begin{aligned}
    0, \quad & f_i^{\text{2D}} < \tau^{\text{2D}}  ~\text{and}~  f_i^{\text{3D}} < \tau^{\text{3D}}, \\ 
    1, \quad & \text{otherwise.}
\end{aligned}
\right.
\label{eq:static_mask}
\vspace{-1mm}
\end{equation}

A pseudo static/dynamic mask $\mathbf{S}_t \in \mathbb{R}^{N_t}$ is produced for the point cloud $\mathbf{P}_t$ at each time frame. Utilizing $\mathbf{S}_t$, $\mathbf{P}_t$ can be separated into two parts: a pseudo dynamic point cloud $\Tilde{\mathbf{P}}_t$ and a pseudo static point cloud $\Bar{\mathbf{P}}_t$. The Chamfer distance loss calculation is then performed on the pseudo dynamic point cloud instead of the entire point cloud. The masked Chamfer loss can be defined as
\begin{equation}
    \mathcal{L}_{mc} = \frac{1}{| \mathbf{T} |} \sum_{t \in \mathbf{T}} \left(\mathcal{L}_{cd} (\Tilde{\mathbf{P}}_t, \Tilde{\mathbf{P}}^{\prime}_t) + \mathcal{L}_{static} (\Bar{\mathbf{P}}_0, t) \right),
\label{eq:mc_loss}
\end{equation}
where $\mathcal{L}_{cd}(\cdot)$ is the Chamfer loss and 
\begin{equation}
    \mathcal{L}_{static} (\Bar{\mathbf{P}}_0, t) = \frac{1}{| \Bar{\mathbf{P}}_0 |} \sum_{p_i \in \Bar{\mathbf{P}}_0} || f_i^t ||_1,
\end{equation}
which pushes the motions of static points to be zero.

\begin{figure}[t]
\centering
\includegraphics[width=1.0\columnwidth]{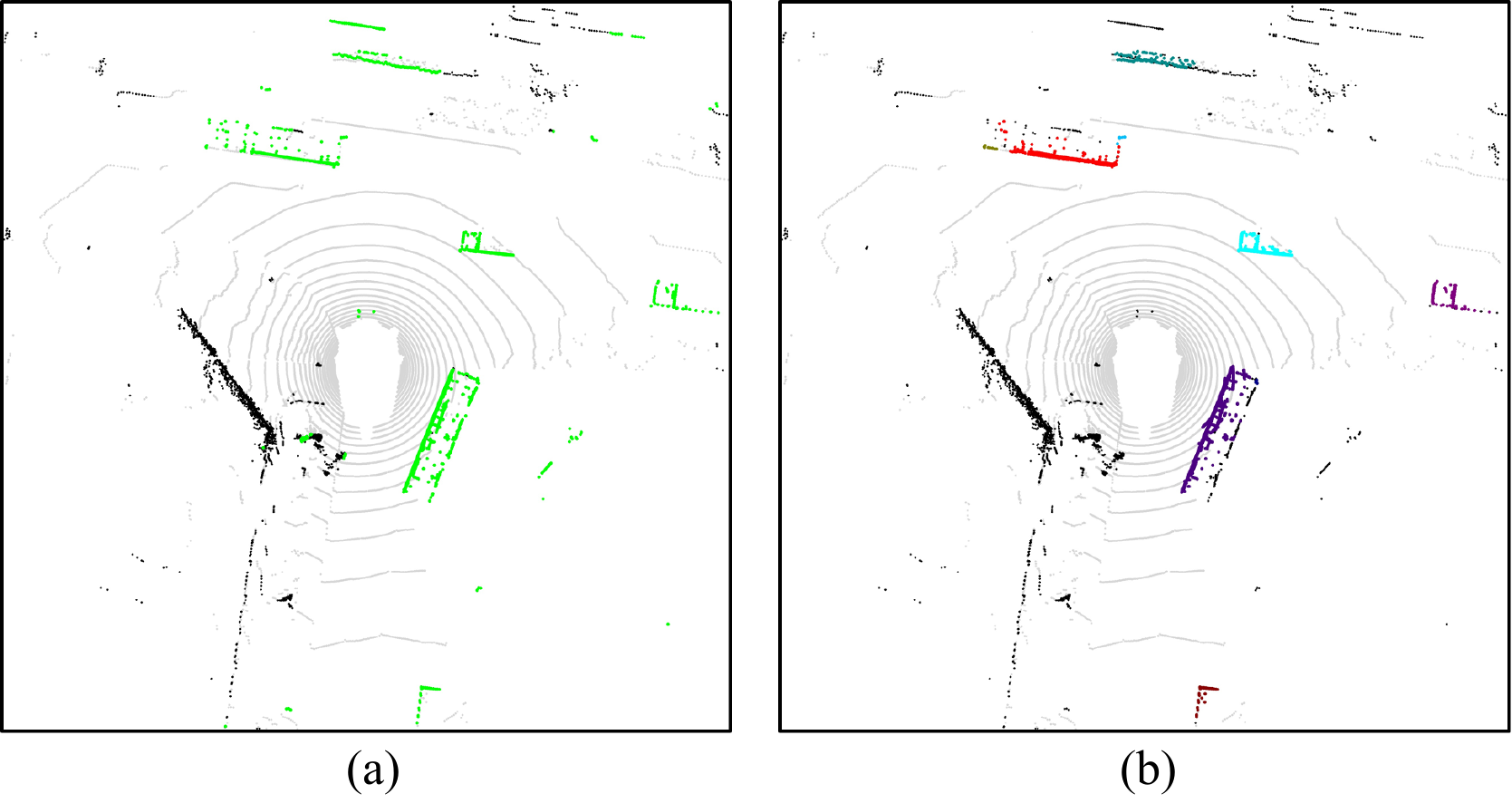} 
\vspace{-3mm}
\caption{An example of the generated static/dynamic mask and the rigid piece labels. Left: green represents dynamic points while black represents static points; Right: each color except black refers to a rigid piece label.}
\label{fig:pseudo_masks}
\vspace{-3mm}
\end{figure}

\subsection{Piecewise Rigidity Loss}

When considering flow estimation on point clouds, local rigidity is an important physical prior that is frequently utilized~\cite{dong2022exploiting, gojcic2021weakly, li2022rigidflow, shen2023self}. Unlike previous self-supervised methods for point clouds that maintain rigidity by clustering on a single-frame point cloud~\cite{li2022rigidflow, shen2023self}, we introduce a simple yet effective method for point cloud piecewise rigidity based on optical flow image clustering.
Compared to a single-frame point cloud, optical flow images exhibit several desirable properties. The motion consistency of dynamic objects is more pronounced in optical flow images and is easier to recognize and extract. In contrast to the sparsity and noise of point clouds, the pixels of moving objects in optical flow images are adjacent, with high uniformity and smoothness. Additionally, the boundaries between objects and the background are more salient.

For the optical flow image $\mathbf{F}^{\text{2D}}$, we apply a simple image clustering method~\cite{achanta2012slic} to obtain over-segmentation labels (Figure \ref{fig:piecewise_rigidity}(b)). The over-segmentation divides the entire image into numerous small pieces. Due to the optical flow consistency of moving objects, it is easy to ensure that almost all pixels in a single small piece belong to the same object.

Through the point cloud to image projection function $\mathcal{T}_{\mathbf{P} \rightarrow \mathbf{I}}$, we can retrieve the corresponding label in the image over-segmentation for each point $p_i$. By projecting the over-segment result labels from $N_c$ camera views, we obtain an over-segmentation result on the whole point cloud.

Some points in the point cloud may be occluded from the camera view due to the slight difference between the LiDAR sensor position and the camera sensor position. For each segment piece, we consider the distance between all points in the piece and the camera and find the smallest distance $d_{min}$. We set a threshold $ \Delta d$ such that if a point in the piece is more than $d_{min} + \Delta d$ away from the camera, it will be excluded from the piece.

As a result, we obtain hundreds of rigid pieces from the point cloud (Figure \ref{fig:piecewise_rigidity}(c)). In order to achieve more accurate and unified segmentation, we employ a simple height-based rigid piece fusion method. Since the visual perspective of the multi-view cameras is typically parallel to the ground, rigid pieces derived from images can easily appear at different heights in the same location. Given the assumption that points within the same grid of the BEV map have the same motion, we consolidate multiple rigid piece labels into a singular label if a grid contains points with distinct labels.

Finally, we generate $N_r$ rigid pieces for the point cloud $\mathbf{P_0}$ (Figure \ref{fig:piecewise_rigidity}(d)). $\mathcal{R}_1, \cdots, \mathcal{R}_{N_r}$ are the $N_r$ rigid pieces, where $\mathcal{R}$ represents a set of points that have the same rigid piece label. For any frame $t \in \mathbf{T}$, the piecewise rigidity loss function is defined as
\begin{equation}
\begin{aligned}
    \mathcal{L}^t_{pr} &= \frac{1}{N_r} \sum_{j=1}^{N_r}  \frac{1}{| \mathcal{R}_j |} \sum_{p_i \in \mathcal{R}_j} | f_{\mathcal{R}_j}^{mean} - f_i^t |,
\end{aligned}
\end{equation}
where $f_{\mathcal{R}_j}^{mean} =\sum_{p_i \in \mathcal{R}_j} f^t_i /| \mathcal{R}_j |, ~~ j=1, \cdots, N_r.$ The final piecewise rigidity loss $\mathcal{L}_{pr}$ is the mean of $\mathcal{L}^t_{pr}$ over all time frames $t$.

\subsection{Temporal Motion Consistency Loss}

For moving objects in traffic scenes, such as cars, pedestrians, and bicycles, their motion patterns do not undergo significant changes over short periods of time. In a point cloud sequence, the displacement of points belonging to a moving object should remain consistent over equal time intervals. Therefore, for self-supervised learning of point cloud sequences, we can apply point-level temporal consistency constraints to the predicted motion. The temporal consistency loss is defined as
\begin{equation}
\begin{aligned}
    \mathcal{L}_{tc} &= \frac{1}{N_0} \sum_{i=1}^{N_0} \frac{1}{|\mathbf{T}|} \sum_{t \in \mathcal{T}} | f_i^{mean} - f_i^t |,
\end{aligned}
\end{equation}
where $f_i^{mean} = \sum_{t \in \mathcal{T}} f_i^t / (t|\mathbf{T}|), ~~ i = 1, \cdots, N_0$.

\subsection{Overall Loss}

In summary, the total loss for the model training is a weighted sum of the proposed masked Chamfer distance loss, piecewise rigidity loss and temporal consistency regularization.

\begin{equation}
    \mathcal{L} = \lambda_{mc} \cdot \mathcal{L}_{mc} + \lambda_{pr} \cdot \mathcal{L}_{pr} + \lambda_{tc} \cdot \mathcal{L}_{tc},
\label{eq:overall_loss}
\end{equation}
where $\lambda_{mc}$, $\lambda_{pr}$, and $\lambda_{tc}$ are the balancing parameters.

%% file: contents/4-experiments.tex
\begin{table*}[t]
\small
\centering
\begin{tabular}{c|c|cc|cc|cc}
\toprule
\multirow{2}{*}{Method} & \multirow{2}{*}{Supervision}  & \multicolumn{2}{c|}{Static} & \multicolumn{2}{c|}{Speed $\leq$ 5 m/s} & \multicolumn{2}{c}{Speed $>$ 5 m/s}  \\
 & & Mean$\downarrow$ & Median$\downarrow$ & Mean$\downarrow$  & Median$\downarrow$ & Mean$\downarrow$ & Median$\downarrow$ \\
\hline
\hline
HPLFlowNet~\cite{gu2019hplflownet}  & supervised  & 0.0041  & 0.0002  & 0.4458  & 0.0960  & 4.3206  & 2.4881 \\
PointRCNN~\cite{shi2019pointrcnn}      & supervised  & 0.0204      & 0             & 0.5514    & 0.1627      & 3.9888    & 1.6252      \\
LSTM-ED~\shortcite{schreiber2019long}        & supervised  & 0.0358      & 0             & 0.3551    & 0.1044      & 1.5885    & 1.0003      \\
MotionNet~\cite{wu2020motionnet}      & supervised  & 0.0201      & 0             & 0.2292    & 0.0952      & 0.9454    & 0.6180      \\
PillarMotion~\cite{pillarmotion}   & supervised  & 0.0245      & 0             & 0.2286    & 0.0930      & 0.7784    & 0.4685      \\
BE-STI~\cite{besti}         & supervised  & 0.0220      & 0             & 0.2115    & 0.0929      & 0.7511    & 0.5413      \\
\midrule
WeakMotionNet~\cite{weakmotionnet}  & weakly sup. & 0.0426      & 0             & 0.4009    & 0.1195      & 2.1342    & 1.2061      \\
\midrule
FlowNet3D~\cite{liu2019flownet3d}      & pre.    & 2.0514      & 0             & 2.2058    & 0.3172      & 9.1923    & 8.4923      \\
HPLFlowNet~\cite{gu2019hplflownet}     & pre.    & 2.2165      & 1.4925        & 1.5477    & 1.1269      & 5.9841    & 4.8553      \\
PillarMotion~\cite{pillarmotion}   & self.        & 0.1620      & 0.0010        & 0.6972    & 0.1758      & 3.5504    & 2.0844      \\
ContrastMotion~\cite{jia2023contrastmotion} & self.        &   \underline{0.0829}      & 0             & \underline{0.4522}    & \textbf{0.0959}      & \underline{3.5266}    & \underline{1.3233}      \\
Ours           & self.        & \textbf{0.0514}           & \textbf{0}             & \textbf{0.4212}         & \underline{0.1073}           & \textbf{2.0766}       & \textbf{1.3226}         \\
\bottomrule
\end{tabular}
\vspace{-1mm}
\caption{Evaluation results of BEV motion prediction on nuScenes~\cite{caesar2020nuscenes} test set. There are four kinds of training supervision: supervised, weakly-supervised (weakly sup.), pre-trained (pre.), and self-supervised (self.). Our method outperforms other self-supervised methods by a significant margin.}
\label{table:main_result}
\vspace{-3mm}
\end{table*}

\section{Experiments}

\subsection{Experimental Setup}

\textbf{Dataset.}
We evaluate our approach on the NuScenes~\cite{caesar2020nuscenes} dataset. NuScenes contains 1000 scenes, each of which has 20 seconds of LiDAR point cloud sequences and multi-view camera videos annotated at 2Hz. Following the setting in previous works for fair comparisons~\cite{wu2020motionnet, besti, pillarmotion, weakmotionnet,jia2023contrastmotion}, we adopt 500 scenes for training, 100 scenes for validation, and 250 scenes for testing. During training, we utilize both the LiDAR point clouds and camera images, while only LiDAR point cloud data is required for the validation and testing of the model. The ground truth BEV motion flow for validation and testing is generated from the detection and tracking annotation from the NuScenes dataset.

\noindent\textbf{Implementation details.}
Initially, the BEV feature maps are extracted from the multi-frame point clouds by \cite{lang2019pointpillars}. Our model backbone is built upon MotionNet~\cite{wu2020motionnet}, which takes sequential BEV features as input and extracts spatial-temporal features. 
The input point clouds are cropped within a range of $[-32,32] \times [-32,32] \times [-3, 2]$ meters, and the BEV output map is $256 \times 256$ in size, which means each cell has a range of $0.25\text{m} \times 0.25\text{m}$.
It is worth noting that our proposed cross-modality self-supervision framework is independent of the network backbone. Also, during the inference process, only sequential point cloud data is needed as the model input.

To generate the optical flow, we employ \cite{teed2020raft} as the optical flow estimation model with the pre-trained parameters offered by Pytorch. The static/dynamic classification thresholds in eq.\ref{eq:static_mask} are $\tau^{\text{2D}} = 5$pixels and $\tau^{\text{3D}} = 1\text{m}$. Besides, we extract the points of the ground plane based on the heights and designate them as the static part of the scene. 
For the training loss in eq.\ref{eq:overall_loss}, we set $\lambda_{mc}=1$, $\lambda_{pr}=0.1$ and $\lambda_{tc}=0.4$. We employ AdamW~\cite{loshchilov2017decoupled} optimization algorithm for training. All models are trained on four NVIDIA 3090 GPUs with a batch size of 64. We train the model for 100 epochs with an initial learning rate of 0.008, and we decay the learning rate by 0.5 every 20 epochs.

\noindent\textbf{Metrics.}
Following previous works~\cite{wu2020motionnet, besti, pillarmotion, weakmotionnet,jia2023contrastmotion}, we use the mean and median errors of motion flow on non-empty cells for evaluation. The error is computed by the L2 distance between the predicted motion flow and ground truth flow for the next 1s future. The final results are presented in three categories divided by varying speeds: static (background and static objects), slow (speed $\leq$ 5 m/s), and fast (speed $>$ 5m/s). 
Regarding the whole model, we directly utilize the 1s future flow output to calculate the metrics. In ablation studies, if the model only predicts the subsequent 0.5s of future flow, we employ linear interpolation to estimate the predicted flow for the next 1s future.

\subsection{Comparison with SOTA Methods}
Table \ref{table:main_result} presents a comprehensive comparison between our proposed self-supervised approach and other methods for BEV motion prediction. Based on the training supervision, all approaches can be categorized into three groups: fully supervised, weakly supervised, and self-supervised. We see that our method achieves state-of-the-art performance in the self-supervised group and surpasses previous methods by a significant margin. Compared to the previous state-of-the-art method~\cite{jia2023contrastmotion}, we exhibit a remarkable improvement of $41\%$ in fast speed metrics, which represent the more challenging and crucial part of motion prediction, $7\%$ in slow speed metrics, and $38\%$ in static metrics. 

~\cite{weakmotionnet} is a weakly supervised method that adopts foreground/background annotation as extra supervision signals. Notably, our method shows comparable performance and even surpasses it in terms of the mean error of fast motion. Furthermore, our method outperforms some fully supervised methods such as \cite{gu2019hplflownet} and \cite{shi2019pointrcnn} by $52\%$ and $48\%$ respectively.

\subsection{Ablation Studies}

\textbf{Masked Chamfer loss.} 
To enhance the robustness of self-supervised learning by mitigating the noises in point cloud sequence data, we design a masked Chamfer distance loss based on the pseudo static/dynamic mask generated from optical flow images. An example of the generated static/dynamic mask is illustrated in Figure \ref{fig:pseudo_masks}(a). Exp. 1\&3, 2\&5 in Table \ref{table:mc_pr_ablation} compare the results of the original Chamfer distance loss (eq.~\ref{eq:cd_loss}) with the masked Chamfer distance loss (eq.~\ref{eq:mc_loss}). We can see that the masked Chamfer distance loss can improve all metrics by a large margin.
Especially for the static motion metrics, the masked Chamfer distance loss can bring up to $75\%$ improvement. This shows its effectiveness of eliminating noise and disturbances originating from the static background, which constitutes the majority of the point cloud data.

\noindent\textbf{Piecewise rigidity.}
To ensure uniformity of motion within the same instance, we design an algorithm to generate instance pieces initially from over-segmentation on optical flow images and propose a piecewise rigidity loss to regulate the motion consistency in each piece. Figure \ref{fig:pseudo_masks}(b) provides an illustration of the generated pieces. 

Exp. 1\&2, 3\&5 in Table \ref{table:mc_pr_ablation} demonstrate the effectiveness of the piecewise rigidity loss, resulting in an improvement of approximately $15\%$ across all evaluation metrics. Exp.4 in Table \ref{table:mc_pr_ablation} utilizes a simple neighborhood smoothness loss to constrain the local rigidity of prediction motion, which serves a similar purpose to our piecewise rigidity approach (see more info in supp.). Exp. 4\&5 indicates that our method outperforms the alternative smoothness loss in performance. Moreover, the piecewise rigidity loss brings significant advantages in terms of training time and computational resources.

\noindent\textbf{Temporal consistency.}
Table \ref{table:temporal_ablation}  presents the results of the ablation study conducted on temporal consistency and prediction frames. It is evident that all experiments incorporating the temporal consistency loss exhibit higher performance, which highlights the effectiveness of temporal consistency as a motion pattern that aids in the learning of motion prediction. 
Furthermore, we explore the impact of different prediction frames on training a motion prediction model. The complete framework predicts the motion of frames -1, 1, and 2 during training with a time interval of 0.5 seconds, and the temporal consistency loss is applied across all predicted frames. In Table \ref{table:temporal_ablation}, the 'past' frame refers to the backward frame -1 and the 'future' frame refers to frame 2. We see that i) As the number of frames involved in motion prediction learning increases, the prediction performance improves correspondingly. This is because the temporal consistency pattern becomes more prominent over a longer point cloud sequence. ii) Predicting backward motion (frame -1) yields a larger improvement compared to predicting a further future frame (frame 2). Due to the ego vehicle's movement, the variations in point cloud data become larger when the time interval expands, which makes learning the correspondence between point clouds a more challenging task

Please refer to the supplementary materials for more qualitative results and ablation studies.

\begin{table}[t]
\small
\begin{tabular}{c|ccc|ccc}
\toprule
\multirow{2}{*}{Exp.} & \multirow{2}{*}{m.c.} & \multirow{2}{*}{smooth.} & \multirow{2}{*}{p.r.} & \multirow{2}{*}{Static} & Speed  & Speed  \\
& & & & & $\leq$ 5 m/s & $>$ 5 m/s \\
\hline
\hline
1&              &                  &                      & 0.2515           & 0.8771         & 3.4098    \\
2&              &                  & \checkmark           & 0.2097           & 0.7135         & 3.1892    \\
3& \checkmark    &                  &                      & 0.0704           & 0.4815         & 2.5389    \\
4& \checkmark    & \checkmark       &                      & 0.0677           & 0.4493         & 2.2142    \\
5& \checkmark    &                  & \checkmark           & 0.0514           & 0.4212         & 2.0766   \\
\bottomrule
\end{tabular}
\vspace{-2mm}
\caption{Ablation of masked Chamfer distance and piecewise rigidity losses. m.c.: masked Chamfer distance loss; smooth.: smoothness regularization; p.r.: piecewise rigidity.}
\label{table:mc_pr_ablation}
\vspace{-2mm}
\end{table}

\begin{table}[t]
\small
\centering
\begin{tabular}{ccc|ccc}
\toprule
\multirow{2}{*}{past} & \multirow{2}{*}{future} & \multirow{2}{*}{temp.} & \multirow{2}{*}{Static} & Speed  & Speed  \\
& & & & $\leq$ 5 m/s & $>$ 5 m/s \\
\hline
\hline
\checkmark    & \checkmark       &                      & 0.1150           & 0.5549         & 2.7503    \\
    &        \checkmark          &             \checkmark         & 0.0748           & 0.5307         & 2.8830    \\
\checkmark    &                  & \checkmark           & 0.0838           & 0.5074         & 2.1814    \\
\checkmark    & \checkmark       & \checkmark           & 0.0514           & 0.4212         & 2.0766   \\
\bottomrule
\end{tabular}
\vspace{-2mm}
\caption{Ablation for prediction frames and temporal consistency loss. past: backward frame -1; future: frame 2 into the future; temp.: temporal consistency loss.}
\label{table:temporal_ablation}
\vspace{-2mm}
\end{table}

%% file: contents/5-conclusion.tex
\vspace{-2mm}
\section{Conclusions}

In this paper, we present a novel cross-modality self-supervised method for BEV motion prediction. 
Concretely, we exploit static/dynamic classification and rigid pieces on point clouds from sequential multi-view images to facilitate motion learning without any manual annotations.
Moreover, we enforce temporal consistency across multiple frames, ensuring temporal smoothness of predicted motion.
Comprehensive experiments conducted on the nuScenes dataset demonstrate that our proposed method achieves state-of-the-art performance and all designed modules are effective.

%% file: contents/acknowledgement.tex
\section{Acknowledgments}

This research is supported by NSFC under Grant 62171276 and the Science and Technology Commission of Shanghai Municipal under Grant 21511100900, 22511106101, and 22DZ2229005.

%% file: contents/supp.tex
\section{Pseudo Static/Dynamic Mask Generation}

To generate the pseudo static/dynamic mask, we categorize each point based on its corresponding optical flow value using a predetermined threshold. Points with optical flow values below the threshold are classified as static. However, employing the identical threshold for the optical flow may not always yield accurate results due to the different distances from points to the camera. Specifically, distant moving objects far from the camera may exhibit small optical flow values and be erroneously classified as static. To address this issue, we utilize a projected 3D scene flow from optical flow to supplement the static assessment, which contains additional spatial information compared to 2D optical flow. We will provide a detailed explanation of how the function $\mathcal{T}_{\text{optf} \rightarrow \text{sf}}$ in eq.4 (in the main paper) is computed.

For point $p_i = (x_i, y_i, z_i) \in \mathbf{P}$, its corresponding image pixel is $(u_i, v_i)$ calculated in eq.2 (in the main paper) by a projection function $\mathcal{T}_{\mathbf{P} \rightarrow \mathbf{I}}$. 

\begin{equation*}
    w_i (u_i, v_i, 1)^{\mathbf{T}} = \mathcal{T}_{\mathbf{P} \rightarrow \mathbf{I}} (x_i, y_i, z_i, 1)^{\mathbf{T}}
\end{equation*}

where $\mathcal{T}_{\mathbf{P} \rightarrow \mathbf{I}} \in \mathbb{R}^{3 \times 4}$ is the projection matrix. Note that in eq.2 (in the main paper), we provide a simplified equation for brevity.

Also, we have calculated the dynamic optical flow $f_i^{\text{2D}} \in \mathbb{R}^2$ after eliminating the effect of ego vehicle motion in eq.3 (in the main paper). Then the end pixel of $f_i^{\text{2D}}$ can be defined as $(u_i^{\prime}, v_i^{\prime}) = (u_i, v_i) + f_i^{\text{2D}}$. 

Since the vertical flow is zero, we can get the endpoint $(x_i^{\prime}, y_i^{\prime}, z_i)$ of $f_i^{\text{3D}}$ by solving the following equation.

\begin{equation*}
    w_i^{\prime} (u_i^{\prime}, v_i^{\prime}, 1)^{\mathbf{T}} = \mathcal{T}_{\mathbf{P} \rightarrow \mathbf{I}} (x_i^{\prime}, y_i^{\prime}, z_i, 1)^{\mathbf{T}}
\end{equation*}

By basic matrix transformation, we have
\begin{equation*}
    (u_i^{\prime}, v_i^{\prime}, 1)^{\mathbf{T}} = \mathcal{T}_{new} (x_i^{\prime} \frac{1}{w_i^{\prime}}, y_i^{\prime} \frac{1}{w_i^{\prime}}, \frac{1}{w_i^{\prime}})^{\mathbf{T}}
\end{equation*}

where $\mathcal{T}_{new} \in \mathbb{R}^{3 \times 3}$ is calculated by 
\begin{equation*}
    \mathcal{T}_{new} = [ \mathcal{T}_{\mathbf{P} \rightarrow \mathbf{I}}[:, :2] ~ | ~ \mathcal{T}_{\mathbf{P} \rightarrow \mathbf{I}}[:, :2]*z_i + \mathcal{T}_{\mathbf{P} \rightarrow \mathbf{I}}[:, :3]]
\end{equation*}

Then $(x_i^{\prime}, y_i^{\prime})$ can be calculated by

\begin{equation*}
    (x_i^{\prime}, y_i^{\prime})^{\mathbf{T}} = \mathcal{T}_{new}^{-1}(u_i^{\prime}, v_i^{\prime}, 1)^{\mathbf{T}}[:2] ~/~ \mathcal{T}_{new}^{-1}(u_i^{\prime}, v_i^{\prime}, 1)^{\mathbf{T}}[2:3]
\end{equation*}

Finally, we have
\begin{equation*}
    f_i^{\text{3D}} = (x_i^{\prime}, y_i^{\prime}, z_i) - (x_i, y_i, z_i)
\end{equation*}

By setting thresholds for both the length values of $f_i^{\text{2D}}$ and $f_i^{\text{3D}}$, the pseudo static/dynamic mask can be finally generated (eq.5 in the main paper).


\section{Piecewise Rigidity Ablation}

In the ablation study regarding piecewise rigidity, we use a smoothness regularization for comparative analysis. We implement the local smoothness loss, which is a widely used technique in scene flow estimation~\cite{wu2019pointpwc, kittenplon2021flowstep3d}, to replace piecewise rigidity. The local smoothness loss can be defined as

\begin{equation*}
    \mathcal{L}_{smooth} = \sum_{p_i \in \mathbf{P}} \frac{1}{\left| \mathcal{N}(p_i) \right|} \sum_{p_k \in \mathcal{N}(p_i)} \left\| f_i - f_k \right\|_2^2
\end{equation*}

where $\mathcal{N}(p_i)$ is the neighbor point set of $p_i$. 

Compared to local smoothness loss, our proposed piecewise rigidity loss has several advantages: 1) The piecewise rigidity ensures instance-level flow uniformity rather than local flow smoothness, aligning more closely with realistic motion pattern; 2) The smoothness regularization can lead to incorrect guidance at the boundaries between dynamic objects and the background; 3) Consequently, our method outperforms smoothness loss, as is shown in the comparison of exp.4 and exp.5 in Table 2 (in the main paper); 4) Our approach offers significantly improved training efficiency. The smoothness loss traverses all the point clouds during computation, while the piecewise rigidity only requires calculations for a few rigid pieces.  

\section{Generated Masks Visualization}
\begin{figure*}[ht]
\centering
\includegraphics[width=0.95\textwidth]{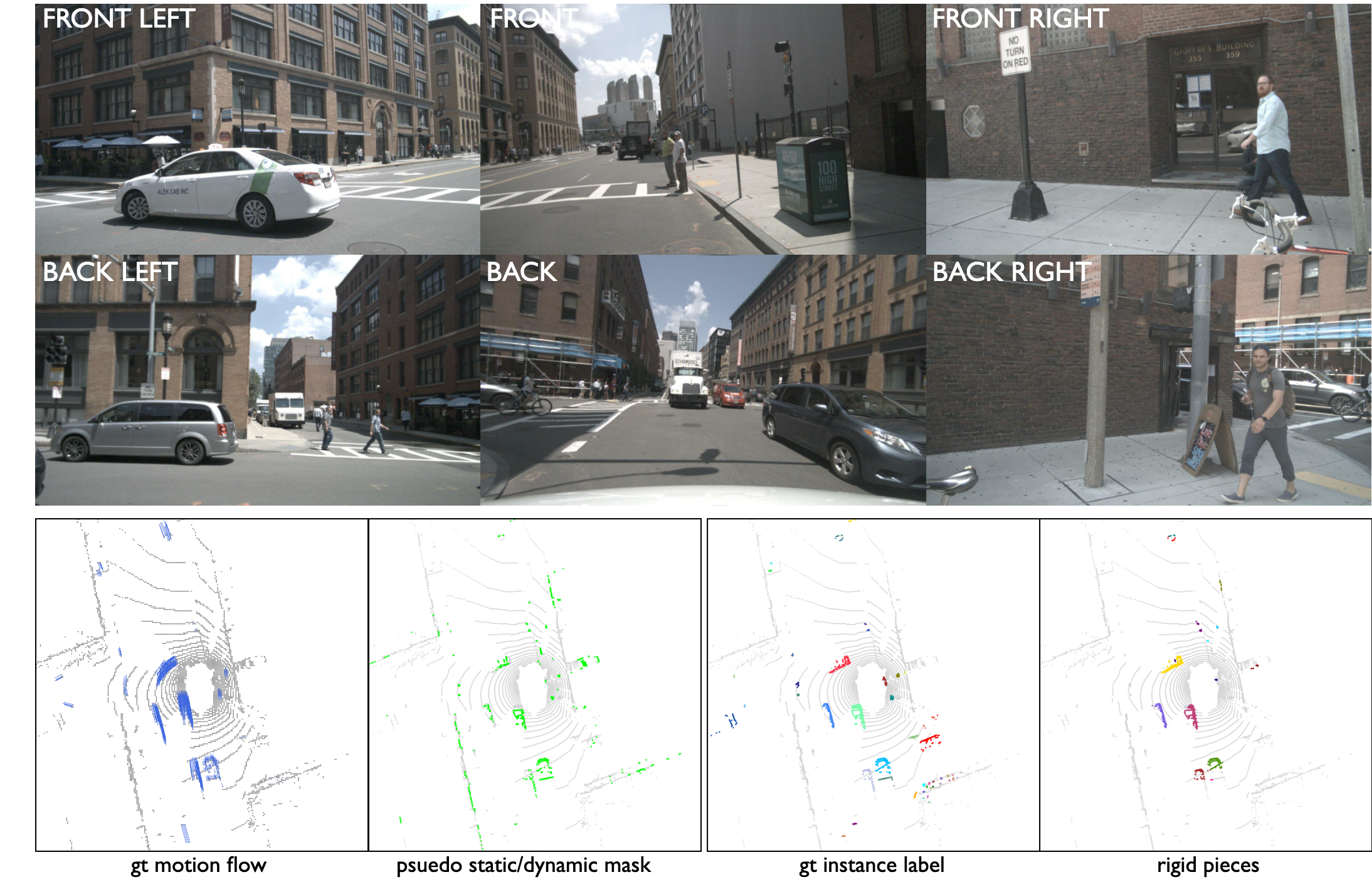}
\vspace{-2mm}
\caption{Visualizations of the pseudo static/dynamic mask and rigid piece labels. A good case.}
\vspace{-2mm}
\label{fig:vis_case1}
\end{figure*}

\begin{figure*}[ht]
\centering
\includegraphics[width=0.95\textwidth]{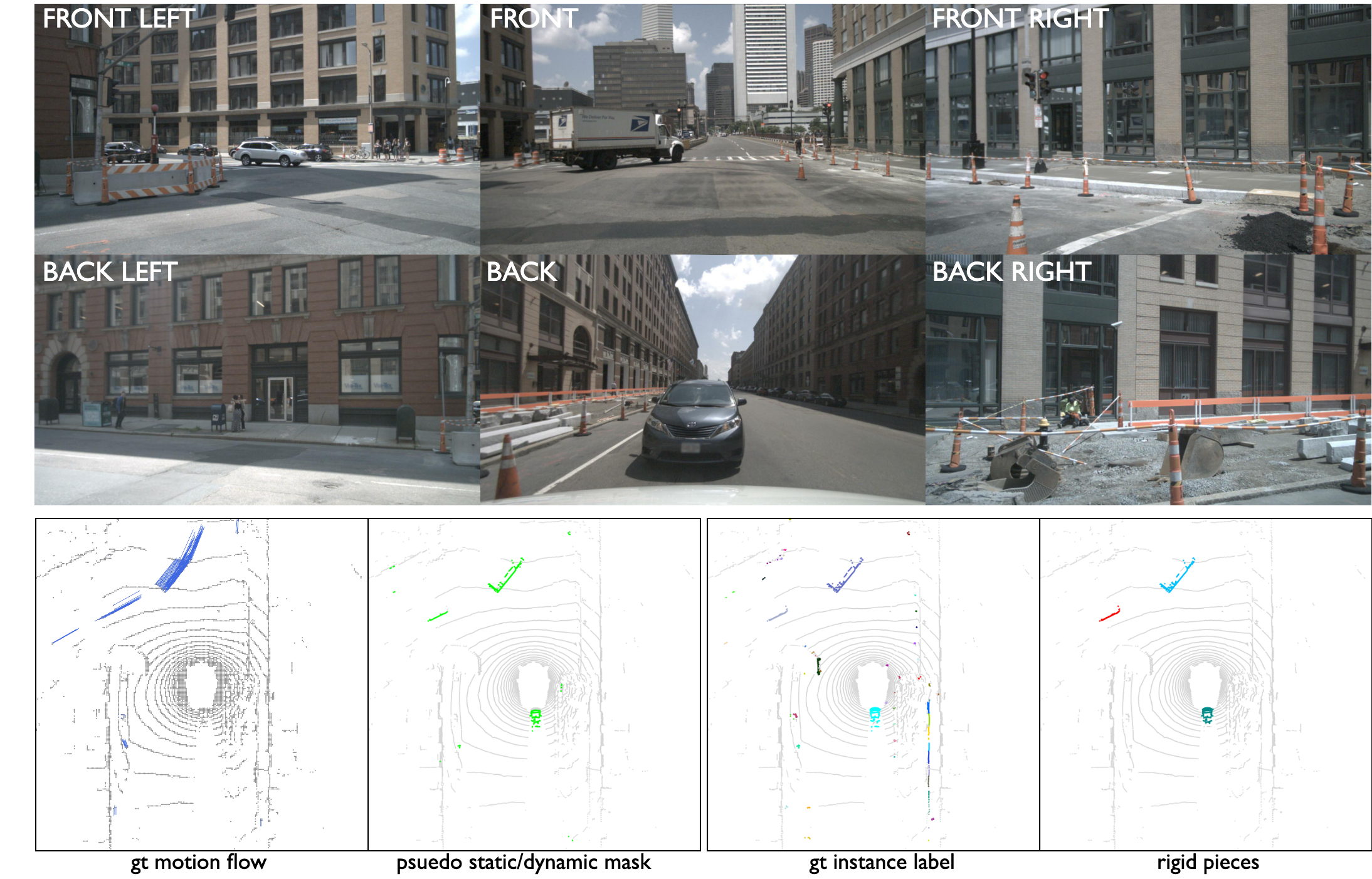}
\vspace{-2mm}
\caption{Visualizations of the pseudo static/dynamic mask and rigid piece labels. A good case.}
\vspace{-2mm}
\label{fig:vis_case2}
\end{figure*}

\begin{figure*}[ht]
\centering
\includegraphics[width=0.95\textwidth]{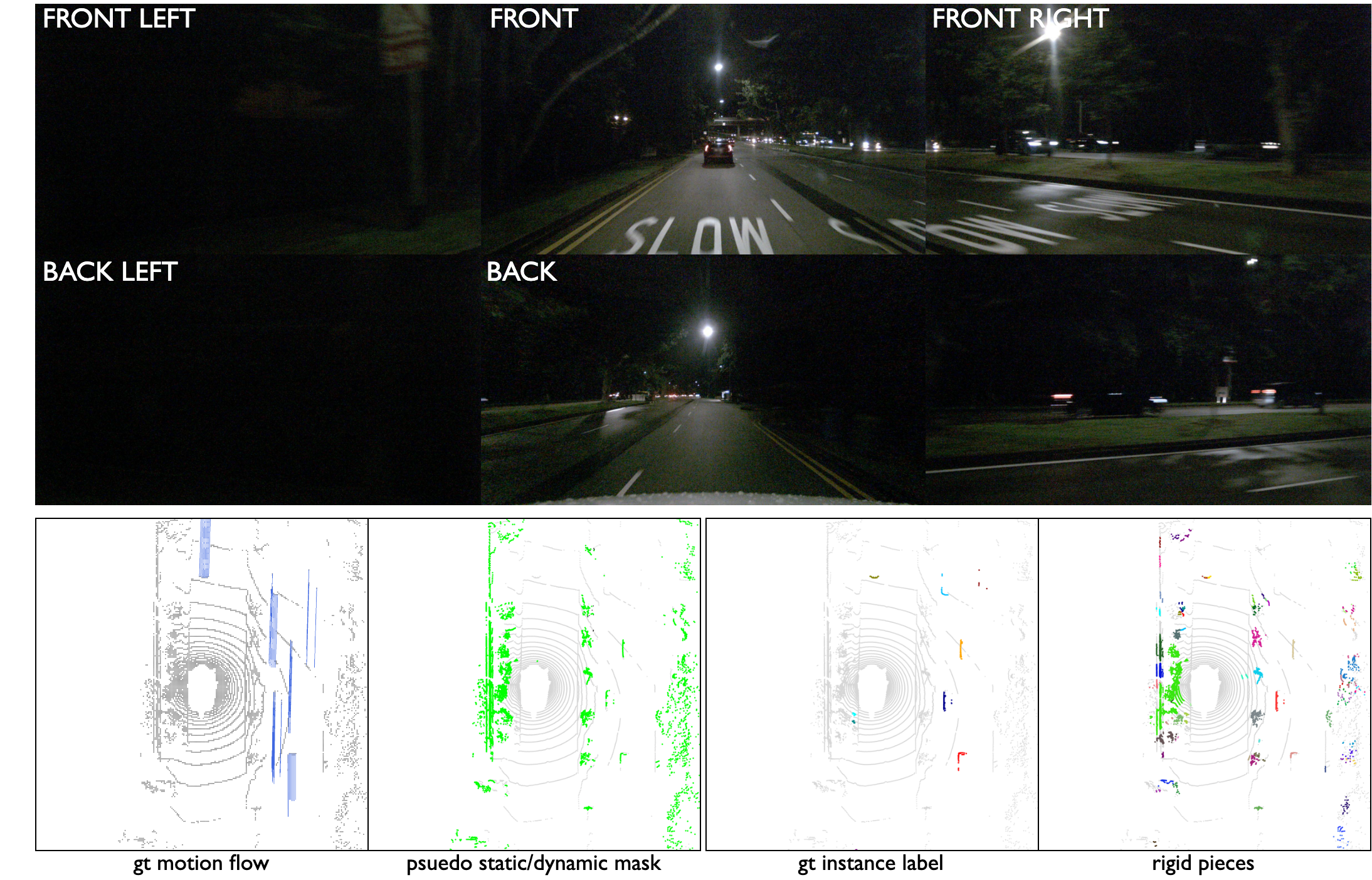}
\vspace{-2mm}
\caption{Visualizations of the pseudo static/dynamic mask and rigid piece labels. A bad case.}
\vspace{-2mm}
\label{fig:vis_case3}
\end{figure*}

We present visualizations of the pseudo static/dynamic mask and rigid piece labels generated by our method under different scenarios. It is evident that the generated labels exhibit high quality in clear weather conditions (Figure \ref{fig:vis_case1} \& \ref{fig:vis_case2}), while the quality is compromised during nighttime periods (Figure \ref{fig:vis_case3}). This discrepancy arises because our label generation heavily relies on camera image data, which is more favorable in good weather conditions. Consequently, our method may have limitations in handling dark or adverse weather data.

Nevertheless, these limitations have a minimal impact on the practicality of our approach. In real-world applications, it is feasible to train the model exclusively using raw data collected during favorable daytime conditions. Since our model's inference solely relies on sequential point cloud inputs, and LiDAR point cloud data is less affected by weather conditions, models trained solely on daytime data can still generalize well to various weather conditions.

It is important to note that, in all our experiments, we did not exclude scenes with nighttime, rainy, or other adverse weather conditions. As a result, data containing noise was also included in the self-supervised training of our model. Introducing weather condition priors and excluding such data can lead to predictable improvements in the performance of our self-supervised learning method.

\section{Argoverse2 Results}

We also conduct experiments on the Argoverse2 dataset~\cite{wilson2023argoverse}, which features diverse sensor configurations with NuScenes~\cite{caesar2020nuscenes}. We extract 12686 samples from the training set for training and 2029/2062 samples from the validation set for validation/testing. The input point clouds are cropped within a range of $[-32,32] \times [-32,32] \times [-3, 2]$ meters, and the BEV output map is also $256 \times 256$ in size, where each cell has a range of $0.25\text{m} \times 0.25\text{m}$.

Table \ref{table:argoverse} shows the results of our method on Argoverse2. Results show that our proposed method is still effective and is comparable to supervised learning. It also demonstrates that our method can be generalized to different scenarios and sensor configurations.

\begin{table}[h]
\scriptsize
\centering
\begin{tabular}{c|ccc}
\toprule
Exp.& Static & Speed $\leq$ 5 m/s  & Speed $>$ 5 m/s  \\
\hline\hline
Zero Flow          &  0.0000       & 0.6104                        & 9.3726      \\
Supervised          &  0.0235       & 0.2310                        & 1.3726      \\
\hline
Ours (w/ m.c. w/o p.r.)    &  0.1419      &       0.4496                  &     2.6137  \\
Ours (w/ m.c. w/ p.r.)    &   \textbf{0.0736}      &     \textbf{0.4221}               &   \textbf{2.3359}   \\
\bottomrule
\end{tabular}
\vspace{-2mm}
\caption{Experimental results on Argoverse2~\cite{wilson2023argoverse}. m.c.: masked Chamfer distance loss; p.r.: piecewise rigidity.}
\label{table:argoverse}
\vspace{-2mm}
\end{table}

\section{Qualitative Comparisons}

\begin{figure*}[ht]
\centering
\includegraphics[width=0.95\textwidth]{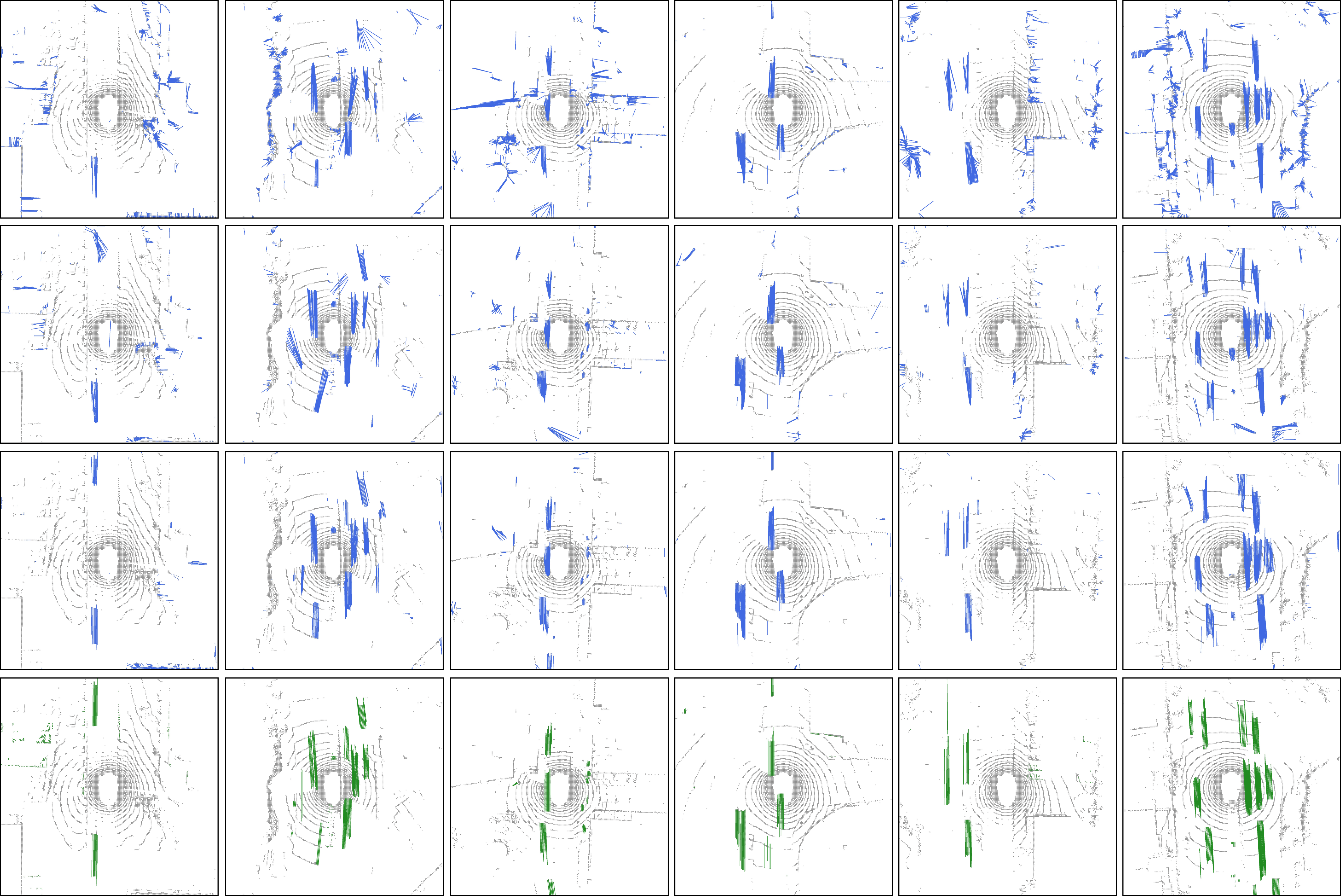}
\vspace{-2mm}
\caption{Qualitative Comparisons. From top to bottom, first row: results without masked Chamfer distance and piecewise rigidity; second row: results without piecewise rigidity; third row: full framework results; fourth row: BEV motion ground truth}
\vspace{-2mm}
\label{fig:vis_result}
\end{figure*}

Figure \ref{fig:vis_result} illustrates the qualitative comparisons for the masked Chamfer distance loss and piecewise rigidity loss. The corresponding quantitative comparisons are in Table 2 in the main paper.

The first row shows the results without both the masked Chamfer distance loss and piecewise rigidity loss. In this case, the model predicts numerous incorrect flows for the background points, primarily due to the presence of substantial noise in the point cloud data.
The second row showcases the results without the piecewise rigidity loss. We can observe that the absence of the piecewise rigidity loss leads to less uniform motion flows for the same object.
Finally, the third row is the results obtained from our full self-supervised training framework.

Overall, these results demonstrate that 1) the masked Chamfer distance design is crucial in preventing the model from predicting incorrect flows for background points affected by noise; 2) the piecewise rigidity loss effectively promotes consistent motion flows for the same object; 3) our self-supervised method produces high-quality motion predictions that closely resemble the ground truth.

%% file: main.bbl
\begin{thebibliography}{50}
\providecommand{\natexlab}[1]{#1}

\bibitem[{Achanta et~al.(2012)Achanta, Shaji, Smith, Lucchi, Fua, and S{\"u}sstrunk}]{achanta2012slic}
Achanta, R.; Shaji, A.; Smith, K.; Lucchi, A.; Fua, P.; and S{\"u}sstrunk, S. 2012.
\newblock SLIC superpixels compared to state-of-the-art superpixel methods.
\newblock \emph{IEEE transactions on pattern analysis and machine intelligence}, 34(11): 2274--2282.

\bibitem[{Baur et~al.(2021)Baur, Emmerichs, Moosmann, Pinggera, Ommer, and Geiger}]{baur2021slim}
Baur, S.~A.; Emmerichs, D.~J.; Moosmann, F.; Pinggera, P.; Ommer, B.; and Geiger, A. 2021.
\newblock Slim: Self-supervised lidar scene flow and motion segmentation.
\newblock In \emph{Proceedings of the IEEE/CVF International Conference on Computer Vision}, 13126--13136.

\bibitem[{Caesar et~al.(2020)Caesar, Bankiti, Lang, Vora, Liong, Xu, Krishnan, Pan, Baldan, and Beijbom}]{caesar2020nuscenes}
Caesar, H.; Bankiti, V.; Lang, A.~H.; Vora, S.; Liong, V.~E.; Xu, Q.; Krishnan, A.; Pan, Y.; Baldan, G.; and Beijbom, O. 2020.
\newblock nuscenes: A multimodal dataset for autonomous driving.
\newblock In \emph{Proceedings of the IEEE/CVF conference on computer vision and pattern recognition}, 11621--11631.

\bibitem[{Casas et~al.(2020)Casas, Gulino, Liao, and Urtasun}]{casas2020spagnn}
Casas, S.; Gulino, C.; Liao, R.; and Urtasun, R. 2020.
\newblock Spagnn: Spatially-aware graph neural networks for relational behavior forecasting from sensor data.
\newblock In \emph{2020 IEEE International Conference on Robotics and Automation (ICRA)}, 9491--9497. IEEE.

\bibitem[{Chen et~al.(2020)Chen, Liu, Feng, Vallespi-Gonzalez, and Wellington}]{chen20203d}
Chen, S.; Liu, B.; Feng, C.; Vallespi-Gonzalez, C.; and Wellington, C. 2020.
\newblock 3d point cloud processing and learning for autonomous driving: Impacting map creation, localization, and perception.
\newblock \emph{IEEE Signal Processing Magazine}, 38(1): 68--86.

\bibitem[{Cheng and Ko(2022)}]{cheng2022bi}
Cheng, W.; and Ko, J.~H. 2022.
\newblock Bi-PointFlowNet: Bidirectional Learning for Point Cloud Based Scene Flow Estimation.
\newblock In \emph{Computer Vision--ECCV 2022: 17th European Conference, Tel Aviv, Israel, October 23--27, 2022, Proceedings, Part XXVIII}, 108--124. Springer.

\bibitem[{Ding et~al.(2023)Ding, Palffy, Gavrila, and Lu}]{ding2023hidden}
Ding, F.; Palffy, A.; Gavrila, D.~M.; and Lu, C.~X. 2023.
\newblock Hidden gems: 4d radar scene flow learning using cross-modal supervision.
\newblock In \emph{Proceedings of the IEEE/CVF Conference on Computer Vision and Pattern Recognition}, 9340--9349.

\bibitem[{Dong et~al.(2022)Dong, Zhang, Li, Sun, and Xiong}]{dong2022exploiting}
Dong, G.; Zhang, Y.; Li, H.; Sun, X.; and Xiong, Z. 2022.
\newblock Exploiting rigidity constraints for lidar scene flow estimation.
\newblock In \emph{Proceedings of the IEEE/CVF Conference on Computer Vision and Pattern Recognition}, 12776--12785.

\bibitem[{Fang et~al.(2023)Fang, Wang, Zhong, Ge, and Chen}]{fang2023tbp}
Fang, S.; Wang, Z.; Zhong, Y.; Ge, J.; and Chen, S. 2023.
\newblock TBP-Former: Learning Temporal Bird's-Eye-View Pyramid for Joint Perception and Prediction in Vision-Centric Autonomous Driving.
\newblock In \emph{Proceedings of the IEEE/CVF Conference on Computer Vision and Pattern Recognition}, 1368--1378.

\bibitem[{Filatov, Rykov, and Murashkin(2020)}]{filatov2020any}
Filatov, A.; Rykov, A.; and Murashkin, V. 2020.
\newblock Any motion detector: Learning class-agnostic scene dynamics from a sequence of lidar point clouds.
\newblock In \emph{2020 IEEE international conference on robotics and automation (ICRA)}, 9498--9504. IEEE.

\bibitem[{Gojcic et~al.(2021)Gojcic, Litany, Wieser, Guibas, and Birdal}]{gojcic2021weakly}
Gojcic, Z.; Litany, O.; Wieser, A.; Guibas, L.~J.; and Birdal, T. 2021.
\newblock Weakly supervised learning of rigid 3D scene flow.
\newblock In \emph{Proceedings of the IEEE/CVF conference on computer vision and pattern recognition}, 5692--5703.

\bibitem[{Gu et~al.(2019)Gu, Wang, Wu, Lee, and Wang}]{gu2019hplflownet}
Gu, X.; Wang, Y.; Wu, C.; Lee, Y.~J.; and Wang, P. 2019.
\newblock Hplflownet: Hierarchical permutohedral lattice flownet for scene flow estimation on large-scale point clouds.
\newblock In \emph{Proceedings of the IEEE/CVF conference on computer vision and pattern recognition}, 3254--3263.

\bibitem[{Hu et~al.(2021)Hu, Murez, Mohan, Dudas, Hawke, Badrinarayanan, Cipolla, and Kendall}]{hu2021fiery}
Hu, A.; Murez, Z.; Mohan, N.; Dudas, S.; Hawke, J.; Badrinarayanan, V.; Cipolla, R.; and Kendall, A. 2021.
\newblock Fiery: Future instance prediction in bird's-eye view from surround monocular cameras.
\newblock In \emph{Proceedings of the IEEE/CVF International Conference on Computer Vision}, 15273--15282.

\bibitem[{Jia et~al.(2023)Jia, Zhou, Zhu, Guo, Zhang, and Ma}]{jia2023contrastmotion}
Jia, X.; Zhou, H.; Zhu, X.; Guo, Y.; Zhang, J.; and Ma, Y. 2023.
\newblock ContrastMotion: Self-supervised Scene Motion Learning for Large-Scale LiDAR Point Clouds.
\newblock \emph{arXiv preprint arXiv:2304.12589}.

\bibitem[{Jund et~al.(2021)Jund, Sweeney, Abdo, Chen, and Shlens}]{jund2021scalable}
Jund, P.; Sweeney, C.; Abdo, N.; Chen, Z.; and Shlens, J. 2021.
\newblock Scalable scene flow from point clouds in the real world.
\newblock \emph{IEEE Robotics and Automation Letters}, 7(2): 1589--1596.

\bibitem[{Khurana et~al.(2023)Khurana, Hu, Held, and Ramanan}]{khurana2023point}
Khurana, T.; Hu, P.; Held, D.; and Ramanan, D. 2023.
\newblock Point Cloud Forecasting as a Proxy for 4D Occupancy Forecasting.
\newblock In \emph{Proceedings of the IEEE/CVF Conference on Computer Vision and Pattern Recognition}, 1116--1124.

\bibitem[{Kittenplon, Eldar, and Raviv(2021)}]{kittenplon2021flowstep3d}
Kittenplon, Y.; Eldar, Y.~C.; and Raviv, D. 2021.
\newblock Flowstep3d: Model unrolling for self-supervised scene flow estimation.
\newblock In \emph{Proceedings of the IEEE/CVF Conference on Computer Vision and Pattern Recognition}, 4114--4123.

\bibitem[{Lang et~al.(2019)Lang, Vora, Caesar, Zhou, Yang, and Beijbom}]{lang2019pointpillars}
Lang, A.~H.; Vora, S.; Caesar, H.; Zhou, L.; Yang, J.; and Beijbom, O. 2019.
\newblock Pointpillars: Fast encoders for object detection from point clouds.
\newblock In \emph{Proceedings of the IEEE/CVF conference on computer vision and pattern recognition}, 12697--12705.

\bibitem[{Lee et~al.(2020)Lee, Kliemann, Gaidon, Li, Fang, Pillai, and Burgard}]{lee2020pillarflow}
Lee, K.-H.; Kliemann, M.; Gaidon, A.; Li, J.; Fang, C.; Pillai, S.; and Burgard, W. 2020.
\newblock Pillarflow: End-to-end birds-eye-view flow estimation for autonomous driving.
\newblock In \emph{2020 IEEE/RSJ International Conference on Intelligent Robots and Systems (IROS)}, 2007--2013. IEEE.

\bibitem[{Li et~al.(2022{\natexlab{a}})Li, Zheng, Li, and Ghanem}]{li2022learning_noisy}
Li, B.; Zheng, C.; Li, G.; and Ghanem, B. 2022{\natexlab{a}}.
\newblock Learning scene flow in 3d point clouds with noisy pseudo labels.
\newblock \emph{arXiv preprint arXiv:2203.12655}.

\bibitem[{Li et~al.(2021)Li, Lin, He, Liu, and Shen}]{li2021hcrf}
Li, R.; Lin, G.; He, T.; Liu, F.; and Shen, C. 2021.
\newblock Hcrf-flow: Scene flow from point clouds with continuous high-order crfs and position-aware flow embedding.
\newblock In \emph{Proceedings of the IEEE/CVF Conference on Computer Vision and Pattern Recognition}, 364--373.

\bibitem[{Li et~al.(2023)Li, Shi, Fu, Wang, and Lin}]{weakmotionnet}
Li, R.; Shi, H.; Fu, Z.; Wang, Z.; and Lin, G. 2023.
\newblock Weakly Supervised Class-Agnostic Motion Prediction for Autonomous Driving.
\newblock In \emph{Proceedings of the IEEE/CVF Conference on Computer Vision and Pattern Recognition}, 17599--17608.

\bibitem[{Li et~al.(2022{\natexlab{b}})Li, Zhang, Lin, Wang, and Shen}]{li2022rigidflow}
Li, R.; Zhang, C.; Lin, G.; Wang, Z.; and Shen, C. 2022{\natexlab{b}}.
\newblock Rigidflow: Self-supervised scene flow learning on point clouds by local rigidity prior.
\newblock In \emph{Proceedings of the IEEE/CVF Conference on Computer Vision and Pattern Recognition}, 16959--16968.

\bibitem[{Li, Kaesemodel~Pontes, and Lucey(2021)}]{li2021neural}
Li, X.; Kaesemodel~Pontes, J.; and Lucey, S. 2021.
\newblock Neural scene flow prior.
\newblock \emph{Advances in Neural Information Processing Systems}, 34: 7838--7851.

\bibitem[{Li et~al.(2022{\natexlab{c}})Li, Yu, Meng, Caine, Ngiam, Peng, Shen, Lu, Zhou, Le et~al.}]{li2022deepfusion}
Li, Y.; Yu, A.~W.; Meng, T.; Caine, B.; Ngiam, J.; Peng, D.; Shen, J.; Lu, Y.; Zhou, D.; Le, Q.~V.; et~al. 2022{\natexlab{c}}.
\newblock Deepfusion: Lidar-camera deep fusion for multi-modal 3d object detection.
\newblock In \emph{Proceedings of the IEEE/CVF Conference on Computer Vision and Pattern Recognition}, 17182--17191.

\bibitem[{Liang et~al.(2022)Liang, Xie, Yu, Xia, Lin, Wang, Tang, Wang, and Tang}]{liang2022bevfusion}
Liang, T.; Xie, H.; Yu, K.; Xia, Z.; Lin, Z.; Wang, Y.; Tang, T.; Wang, B.; and Tang, Z. 2022.
\newblock Bevfusion: A simple and robust lidar-camera fusion framework.
\newblock \emph{Advances in Neural Information Processing Systems}, 35: 10421--10434.

\bibitem[{Liu et~al.(2022)Liu, Lu, Xu, Liu, Li, and Chen}]{liu2022camliflow}
Liu, H.; Lu, T.; Xu, Y.; Liu, J.; Li, W.; and Chen, L. 2022.
\newblock CamLiFlow: Bidirectional camera-LiDAR fusion for joint optical flow and scene flow estimation.
\newblock In \emph{Proceedings of the IEEE/CVF Conference on Computer Vision and Pattern Recognition}, 5791--5801.

\bibitem[{Liu, Qi, and Guibas(2019)}]{liu2019flownet3d}
Liu, X.; Qi, C.~R.; and Guibas, L.~J. 2019.
\newblock Flownet3d: Learning scene flow in 3d point clouds.
\newblock In \emph{Proceedings of the IEEE/CVF Conference on Computer Vision and Pattern Recognition}, 529--537.

\bibitem[{Loshchilov and Hutter(2017)}]{loshchilov2017decoupled}
Loshchilov, I.; and Hutter, F. 2017.
\newblock Decoupled weight decay regularization.
\newblock \emph{arXiv preprint arXiv:1711.05101}.

\bibitem[{Luo, Yang, and Yuille(2021)}]{pillarmotion}
Luo, C.; Yang, X.; and Yuille, A. 2021.
\newblock Self-supervised pillar motion learning for autonomous driving.
\newblock In \emph{Proceedings of the IEEE/CVF Conference on Computer Vision and Pattern Recognition}, 3183--3192.

\bibitem[{Luo, Yang, and Urtasun(2018)}]{luo2018faf}
Luo, W.; Yang, B.; and Urtasun, R. 2018.
\newblock Fast and furious: Real time end-to-end 3d detection, tracking and motion forecasting with a single convolutional net.
\newblock In \emph{Proceedings of the IEEE conference on Computer Vision and Pattern Recognition}, 3569--3577.

\bibitem[{Mittal, Okorn, and Held(2020)}]{mittal2020just}
Mittal, H.; Okorn, B.; and Held, D. 2020.
\newblock Just go with the flow: Self-supervised scene flow estimation.
\newblock In \emph{Proceedings of the IEEE/CVF conference on computer vision and pattern recognition}, 11177--11185.

\bibitem[{Phillips et~al.(2021)Phillips, Martinez, B{\^a}rsan, Casas, Sadat, and Urtasun}]{phillips2021deep}
Phillips, J.; Martinez, J.; B{\^a}rsan, I.~A.; Casas, S.; Sadat, A.; and Urtasun, R. 2021.
\newblock Deep multi-task learning for joint localization, perception, and prediction.
\newblock In \emph{Proceedings of the IEEE/CVF Conference on Computer Vision and Pattern Recognition}, 4679--4689.

\bibitem[{Pontes, Hays, and Lucey(2020)}]{pontes2020scene}
Pontes, J.~K.; Hays, J.; and Lucey, S. 2020.
\newblock Scene flow from point clouds with or without learning.
\newblock In \emph{2020 international conference on 3D vision (3DV)}, 261--270. IEEE.

\bibitem[{Puy, Boulch, and Marlet(2020)}]{puy2020flot}
Puy, G.; Boulch, A.; and Marlet, R. 2020.
\newblock Flot: Scene flow on point clouds guided by optimal transport.
\newblock In \emph{Computer Vision--ECCV 2020: 16th European Conference, Glasgow, UK, August 23--28, 2020, Proceedings, Part XXVIII}, 527--544. Springer.

\bibitem[{Rishav et~al.(2020)Rishav, Battrawy, Schuster, Wasenm{\"u}ller, and Stricker}]{rishav2020deeplidarflow}
Rishav, R.; Battrawy, R.; Schuster, R.; Wasenm{\"u}ller, O.; and Stricker, D. 2020.
\newblock DeepLiDARFlow: A deep learning architecture for scene flow estimation using monocular camera and sparse LiDAR.
\newblock In \emph{2020 IEEE/RSJ International Conference on Intelligent Robots and Systems (IROS)}, 10460--10467. IEEE.

\bibitem[{Schreiber, Hoermann, and Dietmayer(2019)}]{schreiber2019long}
Schreiber, M.; Hoermann, S.; and Dietmayer, K. 2019.
\newblock Long-term occupancy grid prediction using recurrent neural networks.
\newblock In \emph{2019 International Conference on Robotics and Automation (ICRA)}, 9299--9305. IEEE.

\bibitem[{Shen et~al.(2023)Shen, Hui, Xie, and Yang}]{shen2023self}
Shen, Y.; Hui, L.; Xie, J.; and Yang, J. 2023.
\newblock Self-Supervised 3D Scene Flow Estimation Guided by Superpoints.
\newblock In \emph{Proceedings of the IEEE/CVF Conference on Computer Vision and Pattern Recognition}, 5271--5280.

\bibitem[{Shi, Wang, and Li(2019)}]{shi2019pointrcnn}
Shi, S.; Wang, X.; and Li, H. 2019.
\newblock Pointrcnn: 3d object proposal generation and detection from point cloud.
\newblock In \emph{Proceedings of the IEEE/CVF conference on computer vision and pattern recognition}, 770--779.

\bibitem[{Sun et~al.(2018)Sun, Yang, Liu, and Kautz}]{sun2018pwc}
Sun, D.; Yang, X.; Liu, M.-Y.; and Kautz, J. 2018.
\newblock Pwc-net: Cnns for optical flow using pyramid, warping, and cost volume.
\newblock In \emph{Proceedings of the IEEE conference on computer vision and pattern recognition}, 8934--8943.

\bibitem[{Teed and Deng(2020)}]{teed2020raft}
Teed, Z.; and Deng, J. 2020.
\newblock Raft: Recurrent all-pairs field transforms for optical flow.
\newblock In \emph{Computer Vision--ECCV 2020: 16th European Conference, Glasgow, UK, August 23--28, 2020, Proceedings, Part II 16}, 402--419. Springer.

\bibitem[{Tishchenko et~al.(2020)Tishchenko, Lombardi, Oswald, and Pollefeys}]{tishchenko2020self}
Tishchenko, I.; Lombardi, S.; Oswald, M.~R.; and Pollefeys, M. 2020.
\newblock Self-supervised learning of non-rigid residual flow and ego-motion.
\newblock In \emph{2020 international conference on 3D vision (3DV)}, 150--159. IEEE.

\bibitem[{Vora et~al.(2020)Vora, Lang, Helou, and Beijbom}]{vora2020pointpainting}
Vora, S.; Lang, A.~H.; Helou, B.; and Beijbom, O. 2020.
\newblock Pointpainting: Sequential fusion for 3d object detection.
\newblock In \emph{Proceedings of the IEEE/CVF conference on computer vision and pattern recognition}, 4604--4612.

\bibitem[{Wang et~al.(2022)Wang, Pan, Zhu, Wu, Zhan, Jiang, and Yang}]{besti}
Wang, Y.; Pan, H.; Zhu, J.; Wu, Y.-H.; Zhan, X.; Jiang, K.; and Yang, D. 2022.
\newblock Be-sti: Spatial-temporal integrated network for class-agnostic motion prediction with bidirectional enhancement.
\newblock In \emph{Proceedings of the IEEE/CVF Conference on Computer Vision and Pattern Recognition}, 17093--17102.

\bibitem[{Wei et~al.(2023)Wei, Wei, Hu, Lu, Zhong, Chen, and Zhang}]{wei2023asynchrony}
Wei, S.; Wei, Y.; Hu, Y.; Lu, Y.; Zhong, Y.; Chen, S.; and Zhang, Y. 2023.
\newblock Asynchrony-Robust Collaborative Perception via Bird's Eye View Flow.
\newblock In \emph{Thirty-seventh Conference on Neural Information Processing Systems}.

\bibitem[{Wilson et~al.(2023)Wilson, Qi, Agarwal, Lambert, Singh, Khandelwal, Pan, Kumar, Hartnett, Pontes et~al.}]{wilson2023argoverse}
Wilson, B.; Qi, W.; Agarwal, T.; Lambert, J.; Singh, J.; Khandelwal, S.; Pan, B.; Kumar, R.; Hartnett, A.; Pontes, J.~K.; et~al. 2023.
\newblock Argoverse 2: Next generation datasets for self-driving perception and forecasting.
\newblock \emph{arXiv preprint arXiv:2301.00493}.

\bibitem[{Wong et~al.(2020)Wong, Wang, Ren, Liang, and Urtasun}]{wong2020identifying}
Wong, K.; Wang, S.; Ren, M.; Liang, M.; and Urtasun, R. 2020.
\newblock Identifying unknown instances for autonomous driving.
\newblock In \emph{Conference on Robot Learning}, 384--393. PMLR.

\bibitem[{Wu, Chen, and Metaxas(2020)}]{wu2020motionnet}
Wu, P.; Chen, S.; and Metaxas, D.~N. 2020.
\newblock Motionnet: Joint perception and motion prediction for autonomous driving based on bird's eye view maps.
\newblock In \emph{Proceedings of the IEEE/CVF conference on computer vision and pattern recognition}, 11385--11395.

\bibitem[{Wu et~al.(2019)Wu, Wang, Li, Liu, and Fuxin}]{wu2019pointpwc}
Wu, W.; Wang, Z.; Li, Z.; Liu, W.; and Fuxin, L. 2019.
\newblock Pointpwc-net: A coarse-to-fine network for supervised and self-supervised scene flow estimation on 3d point clouds.
\newblock \emph{arXiv preprint arXiv:1911.12408}.

\bibitem[{Zhang et~al.(2022)Zhang, Zhu, Zheng, Huang, Huang, Zhou, and Lu}]{zhang2022beverse}
Zhang, Y.; Zhu, Z.; Zheng, W.; Huang, J.; Huang, G.; Zhou, J.; and Lu, J. 2022.
\newblock Beverse: Unified perception and prediction in birds-eye-view for vision-centric autonomous driving.
\newblock \emph{arXiv preprint arXiv:2205.09743}.

\end{thebibliography}
